\documentclass[10pt,twocolumn,letterpaper]{article}
\usepackage{url}
\usepackage{latex/cvpr}
\usepackage{times}
\usepackage{epsfig}
\usepackage{graphicx}
\usepackage{amsmath}
\usepackage{amssymb}
\usepackage{booktabs}
\usepackage{array, booktabs, makecell}

\cvprfinalcopy 

\def\cvprPaperID{****} 

\begin{document}

\title{Classification for everyone : Building geography agnostic models for fairer recognition}

\author{Akshat Jindal\\
{\tt\small akshatj@stanford.edu}
\and
Shreya Singh\\
{\tt\small ssingh16@stanford.edu}
\and
Soham Gadgil\\
{\tt\small sgadgil@stanford.edu}
}

\maketitle

\begin{abstract}
   In this paper, we analyze different methods to mitigate inherent geographical biases present in state of the art image classification models. We first quantitatively present this bias in two datasets - The Dollar Street Dataset and ImageNet, using images with location information. We then present different methods which can be employed to reduce this bias. Finally, we analyze the effectiveness of the different techniques on making these models more robust to geographical locations of the images.
\end{abstract}

\section{Introduction}
Recent advancements in GPUs and ASICs like TPU, resulting in increased computational power, have led to many object recognition systems achieving state of the art performance on publicly available datasets like ImageNet \cite{5206848}, COCO \cite{lin2014microsoft}, and OpenImages \cite{krasin2017openimages}. However, these systems seem to be biased toward images obtained from well-developed western countries, partly because of the skewed distribution of the geographical source location of such images \cite{de2019does}. As such, these systems do not perform well on images from non-western countries with low income. With such systems being deployed globally for use in many real-world downstream applications, there is a need to make our systems more robust to various geographies. Our project focuses on analyzing domain adaptation techniques to close this performance gap, leading to more robust and fairer object recognition models.

DeVries et al \cite{de2019does} revealed a major gap in the top-5 average accuracy of six object recognition systems on images from high and low income households and images from western and non-western geographies. Our goal is to reduce this bias introduced into the systems because of the inherent nature of the training data. Thus, the task is a simple classification problem, with inputs being the images (resized to 224 $\times$ 224) from two datasets, detailed in section \ref{section:dataset}, and the output being a class prediction. The task would use a simple accuracy based metric for performance measurement, binned according to incomes and geographies as done in \cite{de2019does}.

We fine tune two popular image recognition models to run our experiments -  VGG \cite{simonyan2014very} and ResNet \cite{he2016deep}, both pretrained on ImageNet. First, we test out different techniques to tweak the fine tuning process - weighting the images by income, under/over sampling the images to make the data distribution more uniform, and implementing a focal loss \cite{lin2017focal} function to down-weight the inliers (easy examples) and train on a sparse set of hard examples. We then try Adverserial Discriminative Domain Adaptation (ADDA) \cite{tzeng2017adversarial} to observe if Domain Adaptation is effective in solving the problem at hand.


\section{Related Work}

Our problem setting has been well explored in the analysis by DeVries et al \cite{de2019does}, which analyzes the performance of six object recognition systems on the geographically diverse Dollar Street image dataset \cite{dollar_street}, revealing a major gap in the performance of these systems on images from high and low income households and images from western and non-western geographies. The Dollar Street dataset \cite{dollar_street} is a more geographically diverse dataset when compared to other commonly used image datasets in object recognition. Qualitative analyses suggest that there are two primary reasons behind the disparity in performance across income groups: (1) Visual appearance difference within an object class and (2) Items appearing in different contexts. The authors concluded that there is a need to make object-recognition systems robust across different geographies which motivated our work in this project \cite{10.1007/978-981-10-8639-7_25}.

If we view the problem at hand under the lens of unsupervised domain adaptation, we can leverage the excellent performance of the models on the high income groups to improve the performance on the low income groups \cite{10.1145/3308560.3316599}. There has been extensive work in the field of unsupervised visual domain adaptation. One class of approaches involve the use of domain adversarial objectives whereby a domain classifier is trained to distinguish between the source and target representations while the domain representation is learned so as to maximize the error of the domain classifier. The representation is optimized using the standard minimax objective \cite{ganin2016domain}, or the inverted label objective \cite{tzeng2017adversarial} etc. In this work, we mainly focus on Adversarial Discriminative Domain Adaptation by Tzeng et al\cite{tzeng2017adversarial, 10427646} who use adversarial training to train a target encoder capable of encoding images from the target domain into the same feature space as that of the source images \cite{singh2018footwear}.

We note that there exist many other approaches to domain adaptation notably including a class of approaches which try to learn to convert images from the target domain into the style of the source domain or vice-versa and then use them for classification \cite{singh2019one}. These approaches include works like CycleGANs where the authors combine the standard adversarial loss terms with a cycle-consistency loss term and achieve compelling image-image translation results\cite{zhu2017unpaired, 10390973}. While CycleGAN was not developed for thee purpose of domain adaptation, it gave birth to approaches like CycADA \cite{hoffman2017cycada} where the authors use a slightly modified objective to achieve unsupervised domain adaptation.

\section{Dataset and Features}
\label{section:dataset}
The two datasets we use are The Dollar Street ImageDataset \cite{dollar_street} and ImageNet \cite{5206848}. We also obtain metadata information about the images including the geographic location and income levels.

\subsection{Dollar Street Image Dataset}
The Dollar Street Image Dataset is a collection of $\sim$30000 images taken from over 264 homes in 50 countries. These images belong to 135 classes based on household function and they come along with the location of the image and the income level of the family, adjusted for purchasing power parity. 

The challenge associated with obtaining these images is that there was no public API exposed for trivial image collection. As a result, we had to generate the static html for each of the categorical pages on the website, and then scrape it to get the image url, the location, and the income level. We skip the image urls that gave no response and combine some of the classes which are semantically similar and have few independent samples, to give a final list of 131 classes.

\subsection{ImageNet}
ImageNet is a large scale image database developed for advancing object recognition models. Organized according to the WordNet \cite{miller1998wordnet} hierarchy, ImageNet contains over 14 million hand annotated images spanning more than 20000 categories. 

\begin{figure}[t] 
\begin{center}
   \includegraphics[width=\linewidth]{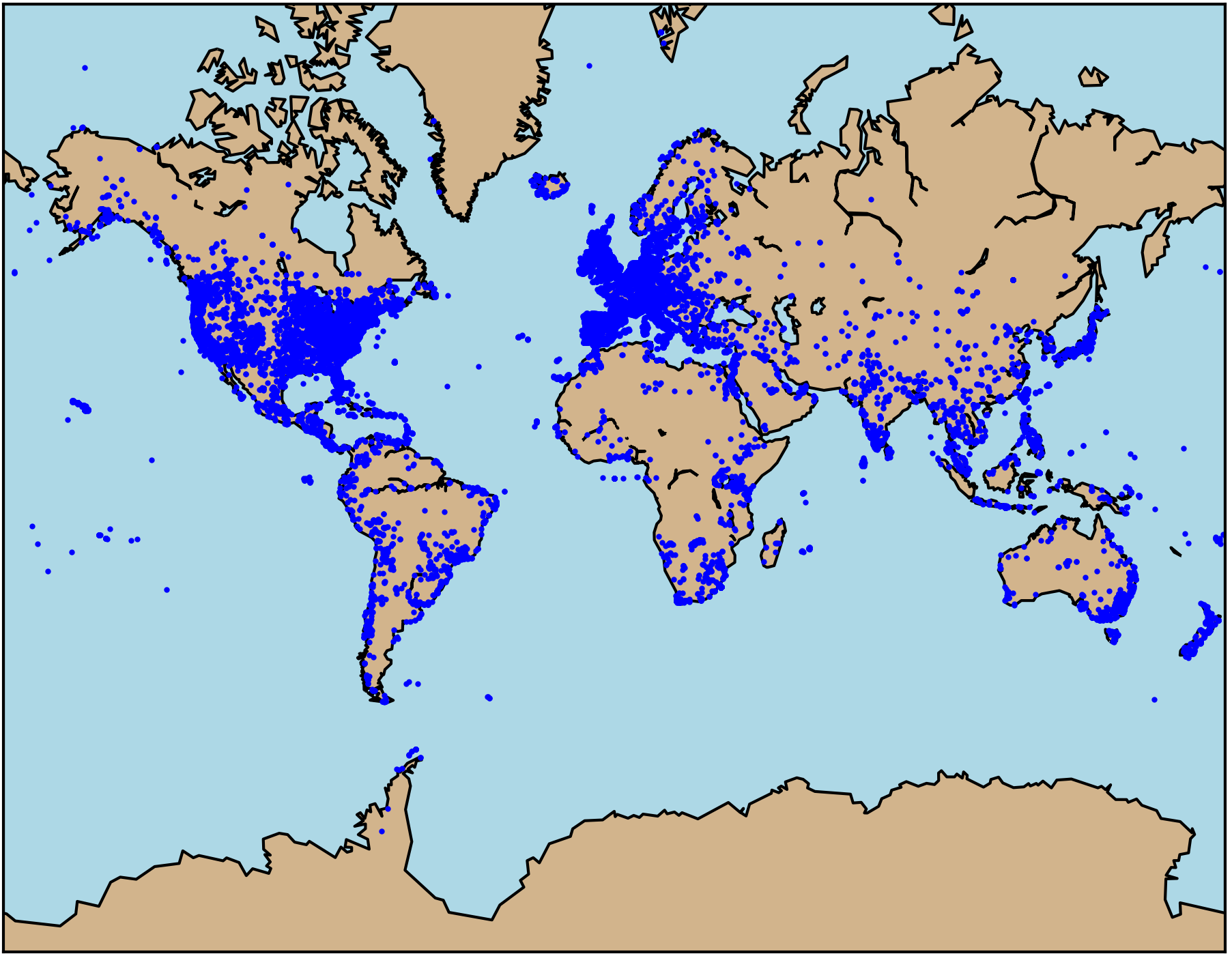}
\end{center}
   \caption{Location Distribution for ImageNet}
\label{fig:img_dist}
\end{figure}

The only images we are interested in are those with geographic location information. These are the images collected from Flickr, which account for $\sim$10-20\% of the entire corpus. Not all the images from Flickr have the location metadata associated with them. Thus, as a preprocessing step, we use the Flickr API to filter images with this metadata available. Because of time constraints, we ran the image collection script for 2 days and were able to obtain 50249 images encompassing 596 classes. This became the core dataset for our experiments with ImageNet. Fig. \ref{fig:img_dist} shows the geographical distribution of the images collected. As we suspected, most of the images are obtained from continents with high income level, such as North America and Europe, while there are few image from continents like Africa and Asia.

Once we had the location information, the next step was to convert these to income levels. For this, we decided to use income levels based on the continent of origin for each image. We used the latitude and longitude of each image to get the continent where it was from and mapped it to the income level using Table \ref{tab:gdp_per_capita}, obtained from Wikipedia \cite{wiki:gdp}. Antartica is not included in the list of continents since its GDP per capita information was not available. This is a very crude mapping for income levels given the extent of the continents, but it does provide us with a ballpark figure.

To the best of our knowledge, there is no publicly available dataset which maps ImageNet images to location and income like we do here. Thus, this dataset can be a contribution in itself.

\begin{table}[t]
    \centering
    \begin{tabular}{l|c}
    \toprule
        \textbf{Continent} & \textbf{GDP Per Capita (US \$)} \\
        \midrule
         Oceania &  53,220 \\
         North America & 49,240 \\ 
         Europe & 29,410 \\
         South America & 8,560 \\ 
         Asia & 7,350 \\
         Africa & 1,930 \\
         \bottomrule
    \end{tabular}
    \vspace{0.5cm}
    \caption{GDP Per Capita (Nominal) by Continent}
    \label{tab:gdp_per_capita}
\end{table}

\section{Methods}
\label{sec:methods}
Our approach is to first observe the extent of income and location biases in popular objection recognition models, VGG \cite{simonyan2014very, 10427781} and ResNet \cite{he2016deep}, pretrained on ImageNet. We then propose \textbf{three} methods: Weighted Loss, Sampling, and Focal Loss for the task of mitigating this bias and achieving fairer results. The same training methodology and fine-tuning architecture is used for both the Dollar Street and the ImageNet dataset.
Additionally, we also explore the Adversarial Discriminative Domain Adaptation \cite{tzeng2017adversarial, 10427909} model as our fourth method and share our results. 

\subsection{Original Models} \label{mod}
To see the performance of the original models to improve upon, we fine-tune the VGG16 and ResNet-18 models (pretrained on ImageNet dataset) on the Dollar Street and ImageNet datasets. This is done by freezing the weights of all layers of the models except for the last one, which we replace with our own custom classifier. The classifier has two fully connected layers with 256 neurons, ReLU activation, and a dropout layer with a dropout probability of 0.3, followed by a softmax output. The models are trained using negative log likelihood loss, Adam optimizer with default hyperparameters, and a batch size of 128. The default learning rate of 1e-3 was chosen after an empirical analysis over the learning rates of 1e-2, 1e-3, and 1e-4.

\subsection{Weighted Loss} \label{weighted_loss}
Our first method for building a geography agnostic model entails a simple income-specific re-weighting of the negative log likelihood loss. 
During training, we take a batch of images and do a forward pass through a VGG16 pre-trained model for the Dollar Street dataset and a ResNet-18 pre-trained model for the ImageNet dataset as defined in Section \ref{mod}. We obtain loss for each training image of the batch and divide it by the income of that training image. For normalization, we also multiply this loss with the mean income of the training batch. Post this, we sum up the individual losses to get a single training loss for the batch and do a backward pass. We outline this more formally through Eq. 1.
\begin{eqnarray}
\mathcal{L}_{batch} =   mean\_batch\_income * \sum_{i=1}^{B}\frac{loss_{i}}{income_{i}}
\end{eqnarray}
Here, $\mathcal{L}_{batch}$ is the weighted batch loss, and $B$ is the batch size. The intuition behind loss re-weighting is to penalize low-income images more so that during training, the classifier learns to classify them better as against the case where there was no re-weighting/penalization.

\subsection{Sampling}
While conducting exploratory data analysis, we observed that the income distribution of training images is highly non-uniform. The consequence of the skewed distribution is the inferior performance of most of the image-classification models which have been pre-trained on high-income images. Our second approach targets this problem before the training phase itself (contrary to the method outlined in Section \ref{weighted_loss} where we tweak the loss during training). We first divide the training images based on their incomes into fix-sized income bins. We then fix an image size threshold of 5000 images and sample (over and under) images from each bin. Specifically, we over sample with replacement and under sample without replacement. We sample images such that the number of images for each income bin is equal to the fixed image size threshold. Doing this will make the income distribution of the training images more uniform which will ensure the model to not be biased toward a particular income group. Post this, we run our original pipeline of image classification as outlined in Section \ref{mod}.

\subsection{Focal Loss}
Class imbalance is a very common problem observed across various datasets which represent real-life scenarios. By definition, class imbalance means that we have an uneven distribution of training samples across different classes. This leads the classification model to give superior results for the majority class(es) and inferior results for the minority class(es). This is primarily because classification models encompassing neural architectures tend to be biased towards the class which is shown to them more often as their weights gets optimized with respect to them more \cite{kumari2017parallelization}. 

In our problem's context, we have a skewed distribution of training images across the income buckets. This leads to the disparity between the low and high income procured images as seen in Fig. \ref{fig:vgg_all} and \ref{fig:imagenet_all}. To tackle this and make the accuracy more uniform across various income buckets, we implemented focal loss \cite{lin2017focal, 10.1145/3610978.3640629} which penalizes easily classified examples more as compared to the difficult examples. The notion of 'penalize' is however, unique in this context. We penalize the easily classified examples by making their contribution to the loss as minimal as possible. Specifically, if there is some example belonging to a particular income bucket (high-income in our case) which is very well classified (probability $>>$ 0.5), then we know it will have a very small loss. However, in a class-imbalanced dataset such as ours, there will be a lot of such examples from a particular income bucket that have small individual loss but when added up, contribute significantly to the final loss. Therefore, to avoid the network being biased to easily classified samples, we multiply their probabilities with an expression involving their softmax normalized scores, thereby diminishing their losses. Formally, we tweak the cross-entropy loss function of a particular sample having softmax score $p_{t}$ as:
\begin{eqnarray}
\mathcal FL(p_{t}) =   -(1-p_{t})^{\gamma}log(p_{t})
\end{eqnarray}
In the above equation, notice that $-log(p_{t})$ is the original cross-entropy loss equation where $p_{t}$ is the softmax score. Here $\gamma$ is a hyperparameter which adjusts the rate at which easy examples are downweighted.


\subsection{ADDA}
\begin{figure*}[t]
\begin{center}
\includegraphics[width = 0.8\linewidth]{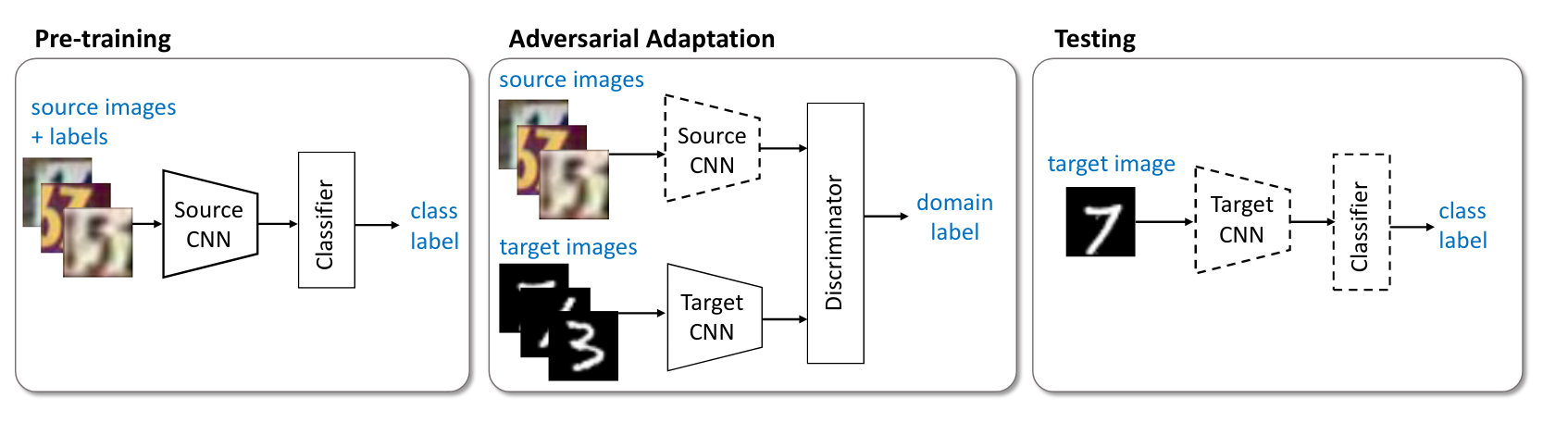}
\end{center}
   \caption{ADDA training procedure. Image taken from \cite{tzeng2017adversarial}}
\label{fig:addatrain}
\end{figure*}

Our third baseline is based on domain adaptation and uses a slightly modified version of the ADDA model \cite{tzeng2017adversarial}. We split the Dollar street dataset \cite{dollar_street} into the source domain which consists of images with income level $> \$600$ and the target domain which consists of low income images with income level $<= \$600 $. The training flow can be best understood from Fig. \ref{fig:addatrain} and consists of the following three steps:
\begin{itemize}
    \item A ResNet-18 model pretrained on ImageNet is used as the source encoder and is finetuned along with the source classifier on the source domain images.
    \item In the adversarial training step, we use a target encoder with the same architecture as the source encoder as the Generator, and a 3 layer FFN as the Discriminator, while holding the source encoder constant. The discriminator's task is to distinguish between the real and fake features from the 2 encoders and the generator's task is to encode the target images into a feature space which is indistinguishable from the feature space of the source encoder.
    \item Finally we test the trained target encoder on the target domain images. The original model simply uses the classifier learned in the first phase to classify the encoded images. Since we have the labels of the target domain, we will fine-tune the classifier on the target domain images as well.
\end{itemize}

\section{Experiments and Results}
In this section, we describe and discuss the results of the experiments run with the techniques mentioned in Section \ref{sec:methods} on the Dollar Street and ImageNet datasets. 

\subsection{Dollar Street Dataset}
\begin{figure}[!ht]
\begin{center}
   \includegraphics[width=\linewidth]{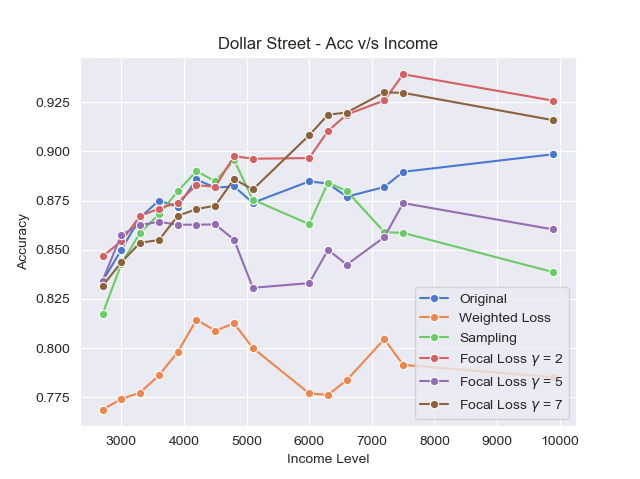}
\end{center}
   \caption{Top-5 Accuracy on Dollar Street v/s Income Level (VGG16)}
\label{fig:vgg_all}
\end{figure}

Table \ref{tab:acc} shows the performance of  different models on the Dollar Street dataset. 
For ADDA, ResNet-18 seems to perform better. Fig. \ref{fig:vgg_all} shows shows the accuracy as a moving average over the preceding ten income buckets as a function of the income levels, divided into buckets of size \$300, on the original model and the four methods mentioned in Section \ref{sec:methods}. Table \ref{tab:adda} shows the top-5 accuracy of the ADDA source encoder on the Dollar Street dataset. Of all the models tested, it seems like the VGG16 model with focal loss ($\gamma = 5$) works the best. It is able to flatten out the accuracy curve across income levels to a reasonable extent, thereby reducing the inherent income-based bias in the pre-trained model.

\begin{table}[t]
\centering 
\begin{tabular}{l|*{2}{c}|*{2}{c}}
  \toprule
    & \multicolumn{2}{c|}{\textbf{VGG16}} & \multicolumn{2}{c}{\textbf{ResNet-18}} \\  
    \midrule
  \textbf{Model} & \thead{Train} & \thead{Val} & \thead{Train} & \thead{Val} \\ 
  \midrule
  Original & 83.28\% & 81.55\% & 84.07\% & 80.56\% \\
  Weighted Loss & 77.71 \% & 75.64 \%  & 80.23\% & 73.41\%\\
  Sampling & 96.05 \% & 77.06\% & 93.88\% & 75.43\%\\
  Focal Loss, $\gamma$=2 & 88.25\% & 81.59\% & 85.13\% & 79.15\% \\ 
  Focal Loss, $\gamma$=5 & 87.83\% & 80.50\% & 83.34\% & 79.86\% \\ 
  Focal Loss, $\gamma$=7 & 87.81\% & 81.28\% & 86.75\% & 79.89\% \\ 
\bottomrule
\end{tabular}
\vspace{1em}
\caption{Top-5 Accuracy on Dollar Street Dataset} 
\label{tab:acc}
\end{table}

\begin{table}[h!]
\centering 
\begin{tabular}{c|*{2}{c}|*{2}{c}}
  \toprule
    & \multicolumn{2}{c|}{\textbf{VGG16}} & \multicolumn{2}{c}{\textbf{ResNet-18}} \\  
    \midrule
  \textbf{Model} & \thead{Train} & \thead{Val} & \thead{Train} & \thead{Val} \\ 
  \midrule
  \thead{Trained on src \\ Eval on src} &   88.27\% &   84.38\% &  90.12\% &  86.37\%  \\
  \thead{Trained on tgt \\ Eval on tgt} &   76.22\% &   72.24\% &  79.26\% &  74.35\%    \\ 
  \thead{Trained on src \\ Eval on tgt} &   88.27\% &   62.28\%&  90.12\%&  63.37\%    \\ 
\bottomrule
\end{tabular}
\vspace{1em}
\caption{Top-5 Accuracy of ADDA Source Encoder on both domains.} 
\label{tab:adda}
\end{table}

\subsubsection{Original Models}
Considering the whole dataset, the models perform well, as shown in Table \ref{tab:acc}. However, the accuracy varies a lot based on the income group of the image. The blue curve in Fig. \ref{fig:vgg_all} shows the accuracy of the original VGG16 model. It is clear that the accuracy of the model increases as the income levels increase, and that there is a substantial gap in the performance of the model on low-income and high-income groups. Our goal would be to make this curve as flat as possible to ensure unbiased performance.

\subsubsection{Weighted Loss}
As show in Table \ref{tab:acc}, weighted loss baseline has an overall train and validation accuracy lower than the original model, however, as shown from Fig. \ref{fig:vgg_all}, it is able to reduce the accuracy deviation. For the original model, there is an increasing trend observed for accuracy as we go from lower-income to higher-income images. Specifically, the accuracy ranges from 83\% to 90 \%. However, with a weighted-loss function, this trend is subdued and we achieve a smaller range through which the accuracy is distributed. This shows that loss re-weighting helps to make the model more robust across different income groups.

\subsubsection{Sampling}
Sampling results in an increase in the overall training dataset. Since, we don't perform sampling over the test set, the ratio of test set images to train set images further reduces. This results in overfitting as now we have a highly augmented train set which leads the model to learn even better on it. The expected large gap between the train and validation accuracy is even seen in Table \ref{tab:acc}. From Fig. \ref{fig:vgg_all}, we observe that the sampling curve closely follows the original curve till income level \$5000. Post that, the sampling curve drops which is a good sign as the accuracy disparity between high and low income groups reduces. Hence, even with sampling, we are able to reduce the accuracy deviation across different income groups, albeit not perfectly.

\subsubsection{Focal Loss}
Replacing the cross-entropy loss with focal loss leads to some interesting results too. Quantitatively, for VGG16, validation accuracy of the original model follidation accuracy of focal loss implemented models. We still see some overfitting in the VGG16 models which could be caused by the focal loss function. This is because, although we hope to tackle the class imbalance problem through our newly implemented focal loss, we have not taken into consideration the scenario of class imbalance (of a slightly different distribution) existing in the validation set. Hence, while training, our model parameters are optimized to be robust across the class distribution of the train set, but might perform worse on the validation set. In Fig. \ref{fig:vgg_all}, focal loss with $\gamma=5$ seems to perform the best across all the models. It gives us accuracy numbers closer to the original model and also reduces the accuracy  deviation across different income groups. It  limits the accuracy numbers between 82.5\% and 87.5\% while in the original model, it is between 82.5\% and 90\%.Initially, we see a downward trend in the accuracy as we increase the income which then picks up later. However, $\gamma = 5$ and $\gamma=7$ give inferior results and increase this disparity even further. On the whole, focal loss with $\gamma=5$ gives accuracy numbers at par with the original model while reducing the accuracy disparity across income groups.

\subsubsection{ADDA}
Preliminary experiments with ADDA are meant to establish a clear domain shift between high income and low income images. The results can be seen in Table \ref{tab:adda} where we can see that a source encoder fine-tuned on the source domain images achieves a validation accuracy of around $87\%$ (on ResNet-18) on the source domain images but if we try to finetune the encoder on low income images (i.e. the target domain), we can only achieve validation accuracy of around $74\%$ (on ResNet-18) on the low income images. Also, we see that if we use an encoder trained on the source domain and evaluate it on the target domain, we get a mere $63\%$ validation accuracy (on Resnet-18). Thus, there is a huge domain shift between low income and high income images.

The experimental results were not as good as expected. They seem to suggest that the domain shift is too large for the model to adapt to leading to the target encoder not being able to learn meaningful representations of the target domain images.

\subsection{ImageNet}
\begin{figure}[!ht]
\begin{center}
   \includegraphics[width=\linewidth]{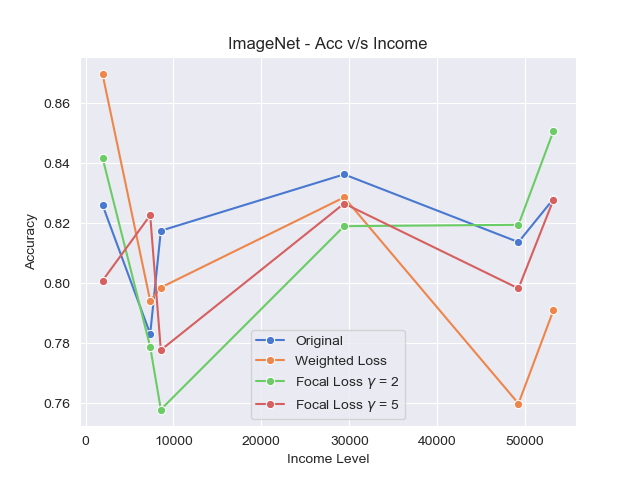}
\end{center}
   \caption{Top-5 Accuracy on ImageNet v/s Income Level (ResNet-18)}
\label{fig:imagenet_all}
\end{figure}

ResNet has been quantitatively proven to perform better than VGG16 on images from ImageNet, based on its win in ILSVRC 2015. Thus, we use ResNet-18 as out pre-trained model to perform experiments on ImageNet. Table \ref{tab:acc_imagenet} shows the performance of ResNet-18 on the ImageNet dataset. The techniques tested here are the ones which performed well on the Dollar Street Dataset. Fig. \ref{fig:imagenet_all} shows the accuracy over the six different income levels corresponding to the six continents. The graph is not as smooth since we have a highly discretized income space. The outcome here does not seem to be as promising as the one from the Dollar Street dataset. However, the model with focal loss ($\gamma = 5$) seems to help mitigate the difference in accuracies to a certain extent.

\begin{table}[t]
\centering 
\begin{tabular}{l|*{2}{c}}
  \toprule
    &  \multicolumn{2}{c}{\textbf{ResNet-18}} \\  
    \midrule
  \textbf{Model} & \thead{Train} & \thead{Val}  \\ 
  \midrule
  Original & 86.66\% & 82\%  \\
  Weighted Loss & 84.51\% & 81.32 \% \\
  Focal Loss, $\gamma$=2 & 84.43\% & 81.4\% \\ 
  Focal Loss, $\gamma$=5 & 87.52\% & 81.17\%\\ 
\bottomrule
\end{tabular}
\vspace{1em}
\caption{Top-5 Accuracy on ImageNet Dataset} 
\label{tab:acc_imagenet}
\end{table}

\subsubsection{Original Models}
The model performs well in terms of accuracy over the entire dataset, as seen in Table \ref{tab:acc_imagenet}. However, as is the case with the dollar street dataset, there is significant variation in the accuracy based on the income level of the location at which the image was taken. Although not as clear as the dollar street dataset, it looks like the accuracy is higher for images from high income level continents, with the highest accuracy being for Europe. The lowest income continent, Africa, seems to have a high accuracy, but that might be because of the high imbalance in the dataset - there are a lot fewer images from Africa than from the other continents. Again, our goal here is to make this curve as flat as possible.
\subsubsection{Weighted Loss}
In Table \ref{tab:acc_imagenet}, the orange line shows the variation of the model with weighted loss with income levels. It seems like this model actually performs worse than the original model; the gap in the accuracies between the low and the high incomes is even larger. This can be because of the way the income reweighting is implemented - low income images are penalized more than high income images. This makes the model focus more on getting the lower income images right, which is evident from the fact that the accuracy increases for the images from lower income levels. However, as a result of this weighting, high income images are penalized less and their accuracy drops significantly. As a result, there is a significant disparity in accuracy observed between the high and low income levels.  
\subsubsection{Focal Loss}
In the models using focal loss, shown by the green and the red lines in Fig. \ref{fig:imagenet_all}, the model with $\gamma$ = 5 seems to have better results than the income reweighted loss. For $\gamma = 2$, the model seems to perform worse than the original model, with a big gap between the accuracies at the low and high income levels. This can be because this value of $\gamma$ doesn't focus on the difficult examples and the easy examples still contribute to most of the model's loss.

However, $\gamma = 5$ seems to present some promising results, although they are not as good as expected. Overall, the accuracy seems to be lower than the original model, but the accuracies for the lowest income level and the income levels in the middle does decrease, resulting in a smaller variance in the accuracy across all the income levels. This indicates that the model has started focusing on the images which are difficult to classify (presumably the images from the lower range of income levels), and these are the images that guide the loss function for most of the training procedure.
\section{Conclusion}
In this project, we reveal the inherent bias in pretrained models like VGG-16 and Resnet-18 leading to huge gaps in performance between high income and low income groups on Dollar Street and Imagenet datasets. We then apply techniques like loss reweighting, sampling, focal loss and explore domain adaptation techniques like ADDA for making these models fairer. We observe that our models work better on Dollar Street than ImageNet which however did seem to give decent results with focal loss with gamma = 5. We explore ADDA and conclude that the domain shift between high and low income groups is too huge to adapt to. For future work, we can have better income information for Imagenet images and also explore more eadvanced domain adaptation techniques like CycADA.

\section{Contributions and Acknowledgements}
All team members contributed equally to the project and Burak Uzkent(SAIL) mentored us through the quarter.

{\small
\bibliographystyle{latex/ieee_fullname}
\bibliography{latex/egbib}
}

\end{document}